\documentclass{article}

\usepackage{PRIMEarxiv}

\usepackage[utf8]{inputenc} 
\usepackage[T1]{fontenc}    
\usepackage[colorlinks=true,linkcolor=blue,citecolor=blue,urlcolor=blue]{hyperref}       
\usepackage{url}            
\usepackage{booktabs}       
\usepackage{amsfonts}       
\usepackage{nicefrac}       
\usepackage{microtype}      
\usepackage{lipsum}
\usepackage{fancyhdr}       
\usepackage{graphicx}       
\graphicspath{{./}}     

\usepackage{multirow}
\usepackage{amsmath,amssymb}
\usepackage{amsthm}
\usepackage{mathrsfs}
\usepackage[title]{appendix}
\usepackage{xcolor}
\usepackage{textcomp}
\usepackage{manyfoot}
\usepackage{algorithm}
\usepackage{algorithmicx}
\usepackage{algpseudocode}
\usepackage{listings}
\usepackage{doi}

\theoremstyle{plain}

\theoremstyle{definition}

\title{A Multi-granularity Concept Sparse Activation and Hierarchical Knowledge Graph Fusion Framework for Rare Disease Diagnosis}

\author{
Mingda Zhang\textsuperscript{1} \and
Na Zhao\textsuperscript{1,2} \and
Jianglong Qin\textsuperscript{1,2}\thanks{Corresponding author: \texttt{qinjianglong@ynu.edu.cn}} \and
Guoyu Ye\textsuperscript{3} \and
Ruixiang Tang\textsuperscript{1} \\[6pt]
\textsuperscript{1}School of Software, Yunnan University, Kunming, China \\
\textsuperscript{2}Yunnan Key Laboratory of Software Engineering, Kunming, China \\
\textsuperscript{3}Yunnan Provincial Hospital of Traditional Chinese Medicine, Kunming, China
}

\begin{document}
\maketitle

\begin{abstract}
Despite advances from medical large language models in healthcare, rare-disease diagnosis remains hampered by insufficient knowledge-representation depth, limited concept understanding, and constrained clinical reasoning. We propose a framework that couples multi-granularity sparse activation of medical concepts with a hierarchical knowledge graph. Four complementary matching algorithms, diversity control, and a five-level fallback strategy enable precise concept activation, while a three-layer knowledge graph (taxonomy, clinical features, instances) provides structured, up-to-date context. Experiments on the BioASQ rare-disease QA set show BLEU gains of 0.09, ROUGE gains of 0.05, and accuracy gains of 0.12, with peak accuracy of 0.89 approaching the 0.90 clinical threshold. Expert evaluation confirms improvements in information quality, reasoning, and professional expression, suggesting our approach shortens the ``diagnostic odyssey'' for rare-disease patients.
\end{abstract}


\section{Introduction}
\label{sec:intro}
Rare diseases affect a small proportion of the population but collectively impact millions worldwide. European standards define rare diseases as having prevalence lower than $5/10{,}000$, while American standards define them as affecting fewer than 200{,}000 individuals~\cite{haendel2020}. Despite their individual rarity, the cumulative patient population is substantial, with approximately 5\% of known rare diseases having available treatments~\cite{tambuyzer2020}.

Patients with rare conditions often experience a prolonged ``diagnostic odyssey,'' requiring 5--7 years from symptom onset to final diagnosis~\cite{wojtara2023}. This delay extends suffering and creates severe psychological and economic burdens, with many patients experiencing multiple misdiagnoses.

Medical large language models (LLMs) show potential in healthcare but face challenges in rare-disease diagnosis, including limited understanding of rare medical concepts, insufficient professional knowledge, restricted reasoning capabilities, and delayed knowledge updates. Schumacher et al.~\cite{schumacher2025} demonstrated that as parameter scale increases, performance improvements follow a logarithmic curve while computational requirements grow exponentially. Current research focuses on enhancing diagnostic capabilities through external knowledge bases while mitigating noise and reasoning inefficiency from traditional dense-retrieval strategies. Despite these advances, Li et al.~\cite{li2023} and Thirunavukarasu et al.~\cite{thirunavukarasu2023} indicate persisting inadequacies in existing models regarding rare-disease expertise.

This study explores medical-concept sparse activation mechanisms and multi-level knowledge-graph fusion to enhance rare-disease diagnostic capabilities. Our main contributions are:
\begin{itemize}
\item \textbf{Multi-granularity sparse activation}: four complementary matching algorithms plus diversity control and fallback mechanisms yield precise, dynamic concept activation while mitigating mis- and missed diagnoses.
\item \textbf{Three-layer knowledge graph with real-time updates}: taxonomy--clinical--instance layers fused with web search provide flexible representation and fresh evidence.
\end{itemize}

\section{Related Work}
\label{sec:related}

\subsection{Applications and Challenges of LLMs in Rare-disease Diagnosis}
LLMs acquire extensive knowledge through pre-training but exhibit significant gaps in rare disease diagnosis. Current medical LLMs primarily incorporate medical knowledge through pre-training on large medical datasets~\cite{wu2020}. Knowledge enhancement methods include retrieval-augmented generation, external tool calling, and parameter fine-tuning, each with different advantages. Chen et al.~\cite{chen2024} identified data scarcity, delayed knowledge updates, and complex reasoning chains as primary challenges.

Fine-tuning LLMs on domain-specific corpora can improve recognition capabilities for rare disease concepts~\cite{wang2023}. This research combines retrieval-augmented generation with knowledge graph fusion, integrating structured medical knowledge with real-time web search results to address both comprehensiveness and timeliness challenges.

\subsection{Concept Sparse Activation Mechanisms}
The concept sparse activation mechanism was initially proposed by Wang et al.~\cite{wang2022} but faces challenges when applied to rare disease diagnosis. Traditional diagnostic methods rely on manual expert screening, while modern natural language processing tools can extract rare disease information from clinical texts~\cite{yang2023}. Traditional sparse activation methods have limitations in handling the complexity and diversity of rare disease terminology. Chen et al.~\cite{chen2024} proposed a knowledge integration framework but failed to address diverse linguistic expressions. Khoshnevisan et al.~\cite{khoshnevisan2024} explored context-aware concept activation methods but lacked optimization for special scenarios of rare diseases.

Hybrid frameworks integrating dictionary-based natural language processing tools with LLMs can improve rare disease recognition accuracy~\cite{huang2024}, demonstrating advantages in handling complex medical terminology variants and low-resource language expressions~\cite{young2025}. The multi-granularity activation mechanism proposed in this study combines four complementary matching strategies with diversity control and fallback mechanisms to form a comprehensive recognition system.

\subsection{Medical Knowledge Graphs}
Medical knowledge graphs face technical challenges in rare disease diagnosis, particularly in knowledge representation granularity, timeliness, and complex relationship expression~\cite{wu2020}. Current approaches often fail to effectively integrate real-time web knowledge update mechanisms, limiting their practicality in rapidly evolving research fields. Zhu et al.~\cite{zhu2020a} and Wu et al.~\cite{wu2023} explored the application potential of medical knowledge graphs in rare disease diagnosis but failed to effectively integrate real-time web knowledge update mechanisms.

The three-layer architectural knowledge graph designed in this study divides knowledge into classification ontology, clinical feature, and instance layers, connecting abstract concepts with specific clinical cases. This structure expresses complex relationships between rare disease concepts and addresses timeliness through regular updates combined with real-time retrieval. The integration of natural disease process data and real-world data has important value for comprehensively describing disease progression and discovering novel biomarkers~\cite{subramanian2020}.

\section{Core Technologies and Implementation}
\label{sec:methods}

\subsection{Multi-granularity Medical Concept Sparse Activation Mechanism}
\begin{figure}
\centering
\includegraphics[width=0.9\textwidth]{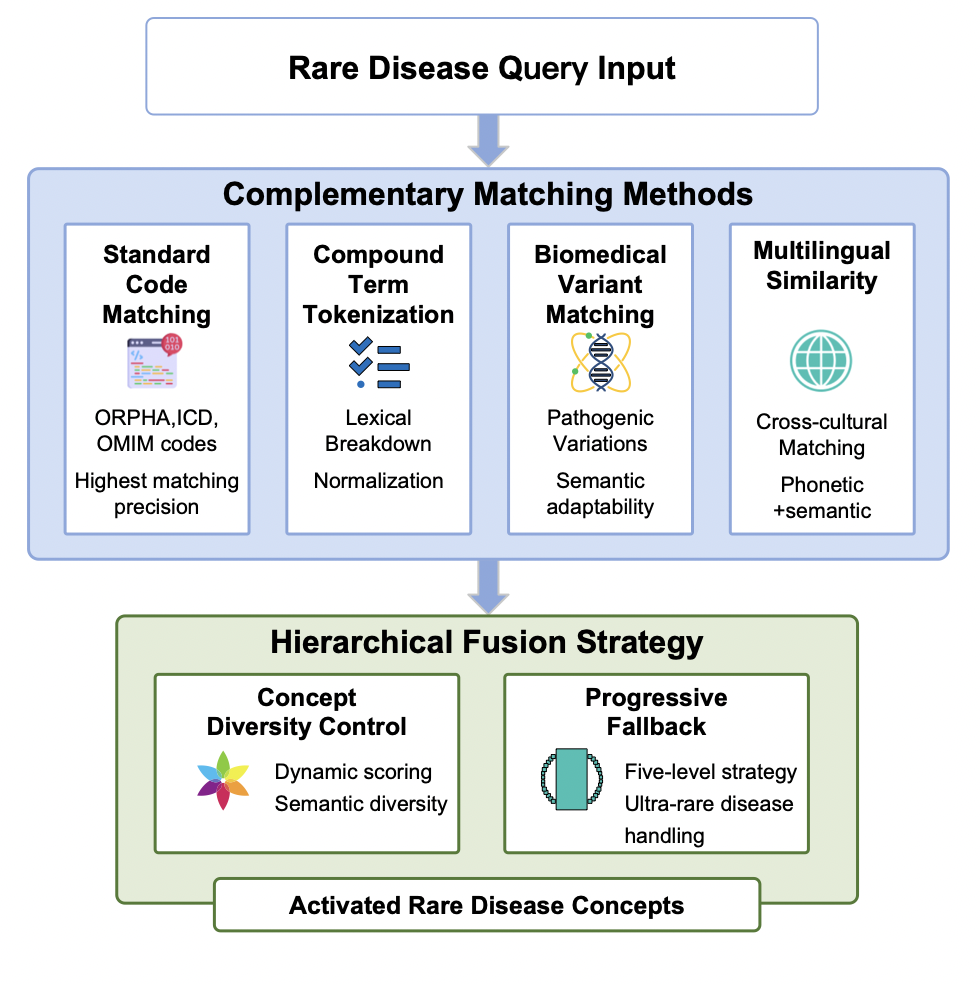}
\caption{Multi-granularity Medical Concept Sparse Activation Diagram for Rare Diseases}
\label{fig:concept_activation}
\end{figure}

The multi-granularity medical concept sparse activation mechanism for rare diseases proposed in this research constitutes the core framework for diagnosing rare diseases. This framework, as shown in Figure~\ref{fig:concept_activation}, consists of three core functional modules: four complementary matching methods, rare disease concept diversity control, and a five-level progressive fallback strategy. In the complementary matching methods module, standardized coding matching precisely identifies standard codes such as ORPHA, International Classification of Diseases, and Online Mendelian Inheritance in Man and ensures the highest matching weight; compound terminology segmentation performs text standardization processing through medical terminology segmentation; biomedical variant matching extends to pathogen variants and effectively handles semantic drift; multilingual cross-cultural similarity processing addresses multilingual cultural environments and comprehensively utilizes phonetic n-gram and semantic embedding technologies. Among these, the rare disease concept diversity control mechanism ensures semantic diversity coverage through dynamic score adjustment, while the five-level progressive fallback strategy specifically addresses no-match situations for ultra-rare diseases, collectively forming a comprehensive rare disease medical concept recognition framework.

\subsubsection{Mathematical Model of Rare Disease Multilingual Cross-cultural Matching Algorithm}

To achieve high-precision rare disease concept recognition, this research designed a series of complementary matching algorithms. The rare disease standardized coding matching algorithm is defined as:
\begin{equation}
M_{\text{code}}(T_s, T_t) = \begin{cases}
1.0, & \text{if } ID(T_s) = ID(T_t) \\
0.8, & \text{if } S_N(T_s) = S_N(T_t) \\
\sum_{i=1}^{n} w_i \cdot \delta(ALIAS_i(T_s), ALIAS_i(T_t)), & \text{otherwise}
\end{cases}
\end{equation}
where $M_{\text{code}}(T_{s},T_{t})$ represents the matching score function with output range $[0,1]$, $T_{s}$ and $T_{t}$ refer to rare disease terminology in source and target languages, $ID(T)$ is the standardized identifier extraction function (ORPHA, ICD-10/11, OMIM), $S_N(T)$ denotes the standard name retrieval function, $ALIAS_{i}(T)$ represents the official alias retrieval function, $w_{i}$ indicates the weight coefficient for each alias based on source authority, and $\delta(a,b)$ is an indicator function that equals 1 when $a=b$ and 0 otherwise.

The rare disease compound terminology word segmentation matching algorithm is:
\begin{equation}
M_{\text{term}}(T_{s},T_{t}) = \frac{\sum_{i = 1}^{m}\sum_{j = 1}^{n}c(S_{i},T_{j}) \cdot w(S_{i}) \cdot w(T_{j})}
{\sum_{i = 1}^{m}w(S_{i}) \cdot \sum_{j = 1}^{n}w(T_{j})}
\end{equation}
where $M_{\text{term}}(T_{s},T_{t})$ represents the matching score function with a range of $[0,1]$, $S_{i}$ and $T_{j}$ are semantic units after word segmentation, $c(S_{i},T_{j})$ denotes the matching degree between semantic units with values ranging from $[0,1]$, and $w(S_{i})$ and $w(T_{j})$ refer to the weight coefficients assigned to semantic units.

The rare disease biomedical variant matching algorithm is:
\begin{equation}
M_{\text{var}}(T_{s},T_{t}) = \max\{Sim_{\text{abbr}}(T_{s},T_{t}),Sim_{\text{part}}(T_{s},T_{t}),Sim_{\text{sem}}(T_{s},T_{t})\}
\end{equation}
where $Sim_{\text{abbr}}$ represents the abbreviation similarity function, $Sim_{\text{part}}$ denotes the partial matching similarity function, and $Sim_{\text{sem}}$ is the semantic equivalence similarity function.

The rare disease multilingual cross-cultural similarity algorithm is:
\begin{equation}
M_{\text{multi}}(T_{s},T_{t}) = f_{\text{combine}}(Sim_{\text{trans}}(T_{s},T_{t}),Sim_{\text{char}}(T_{s},T_{t}),Sim_{\text{emb}}(T_{s},T_{t}))
\end{equation}
where $Sim_{\text{trans}}$ refers to the transliteration similarity function, $Sim_{\text{char}}$ is the character sequence similarity function, $Sim_{\text{emb}}$ represents the semantic embedding similarity function, and $f_{\text{combine}}$ denotes the combination function that integrates the three similarity measures.

\subsubsection{Rare Disease Concept Diversity Control Mechanism}
Traditional activation methods may cause misdiagnosis by focusing on singular perspectives. Our concept diversity control mechanism dynamically adjusts activation scores:
\begin{equation}
S_{\text{final}}^{\prime}(c, q) = \begin{cases} 
\lambda_{\text{RD}} \times S_{\text{final}}(c, q), & \text{if } c \in C_{\text{used}} \\ 
S_{\text{final}}(c, q), & \text{otherwise} 
\end{cases}
\end{equation}
where $\lambda_{\text{RD}}$ is the diversity control factor, $C_{\text{used}}$ represents the set of historically activated concepts, $S_{\text{final}}(c, q)$ indicates the original concept activation score, and $S_{\text{final}}^{\prime}(c, q)$ denotes the adjusted concept activation score.

The concept diversity evaluation metric is:
\begin{equation}
D(C_{\text{active}}) = 1 - \frac{|C_{\text{active}} \cap C_{\text{used}}|}{|C_{\text{active}}|}
\end{equation}
where $D(C_{\text{active}})$ represents the diversity score ranging from $[0,1]$, $C_{\text{active}}$ is the set of currently activated concepts, and $C_{\text{used}}$ denotes the set of previously used concepts.

\subsubsection{Specialized No-Match Concept Fallback Mechanism for Rare Diseases}
For ultra-rare diseases where traditional activation fails, we developed a hierarchical fallback mechanism:

This mechanism employs five progressive levels: Level 1 involves same-family rare disease fallback; Level 2 utilizes phenotype-driven fallback using Human Phenotype Ontology; Level 3 implements clinical feature combination fallback; Level 4 applies genotype association fallback; and Level 5 employs rare disease basic knowledge fallback.

\subsubsection{Adaptive Sparse Control for Rare Diseases}
For complex rare disease diagnostic queries, our adaptive sparse control strategy is:
\begin{equation}
k_{\text{RD}} = \max\{k_{\text{min}}^{\text{RD}},\min(k_{\text{max}}^{\text{RD}}, \alpha_{\text{RD}} \times |C_{\text{RD}}| \times C_{\text{RD}}(q))\}
\end{equation}
where $k_{\text{RD}}$ represents the number of concepts to activate, $k_{\text{min}}^{\text{RD}}$ and $k_{\text{max}}^{\text{RD}}$ denote the minimum and maximum concept numbers respectively, $\alpha_{\text{RD}}$ is the basic sparsity parameter, $|C_{\text{RD}}|$ indicates the total number of concepts in the knowledge base, and $C_{\text{RD}}(q)$ refers to the query complexity evaluation function.

The rare disease query complexity evaluation function is:
\begin{equation}
C_{\text{RD}}(q) = \beta_{\text{RD}}^{1} \times L(q) + \beta_{\text{RD}}^{2} \times T_{\text{RD}}(q) + \beta_{\text{RD}}^{3} \times S_{\text{RD}}(q) + \beta_{\text{RD}}^{4} \times M_{\text{RD}}(q)
\end{equation}
where $L(q)$ represents the query length factor, $T_{\text{RD}}(q)$ denotes the terminology density factor, $S_{\text{RD}}(q)$ refers to the semantic complexity factor, $M_{\text{RD}}(q)$ indicates the multi-system manifestation factor, and $\beta_{\text{RD}}^{1-4}$ are the weight coefficients determined through regression analysis.

\subsection{Three-layer Medical Knowledge Graph Architecture and Web Search Integration}
\subsubsection{Multi-level Knowledge Structure Design for Rare Diseases}
\begin{figure}
\centering
\includegraphics[width=0.9\textwidth]{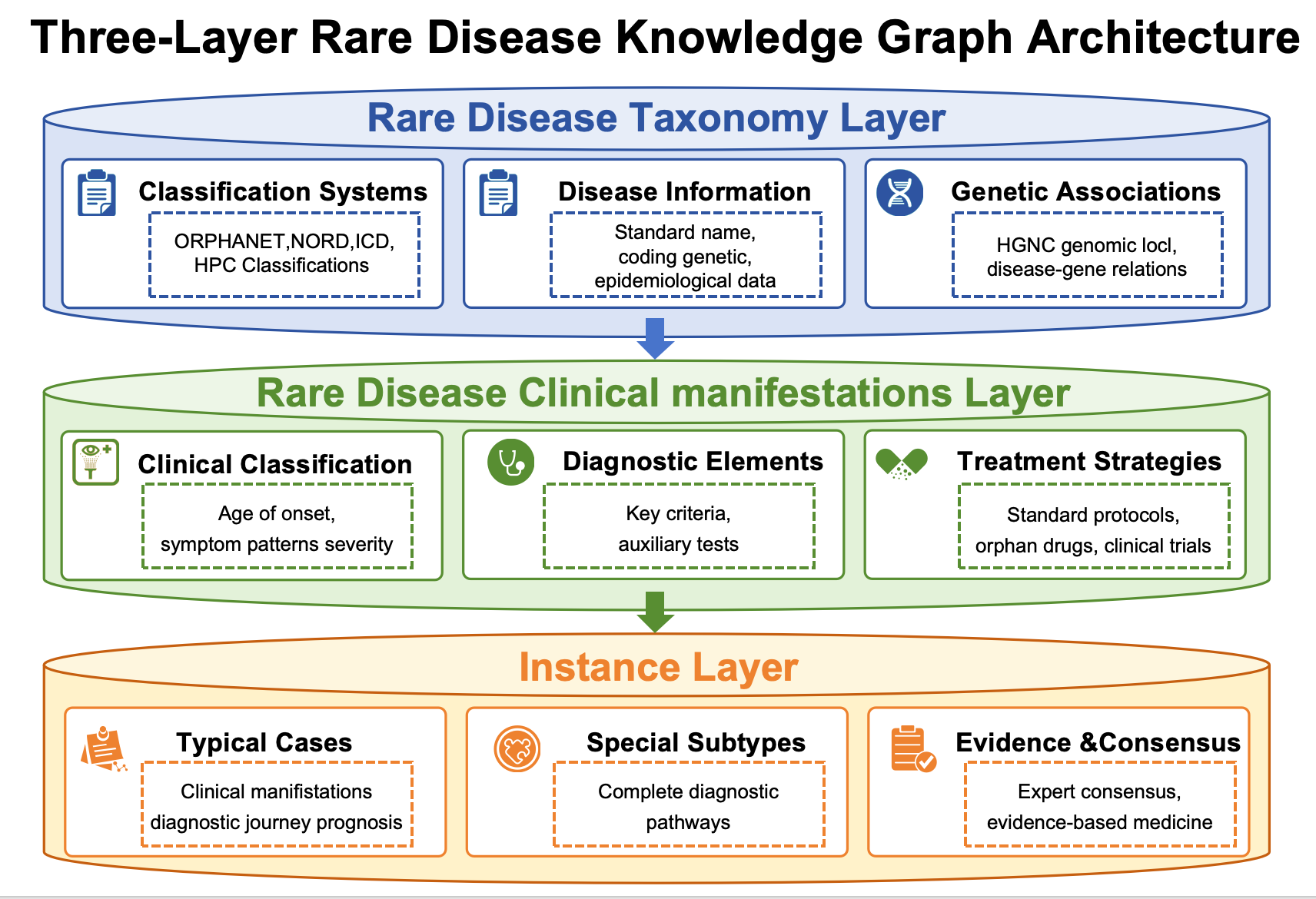}
\caption{Three-layer Medical Knowledge Graph Architecture}
\label{fig:knowledge_graph}
\end{figure}

Our specialized three-layer medical knowledge graph architecture is shown in Figure~\ref{fig:knowledge_graph}.

The architecture consists of three main components: the Classification Layer, which integrates authoritative systems like ORPHANET, NORD, ICD, and HPO; the Clinical Manifestation Layer, which elaborates disease phenotypic features and diagnostic criteria; and the Instance Layer, which collects typical cases, special subtypes, and diagnostic pathways.

\subsubsection{Integration of Knowledge Graph and Web Search}
The system integrates multiple APIs and specialized search engines, constructing a multi-source retrieval network with adaptive query strategies optimized for rare diseases. This integration solves the delayed update problem of rare disease knowledge, providing comprehensive and up-to-date support for large language models.

\section{Experiments and Evaluation}
\label{sec:experiments}

\subsection{Experimental Setup}
To evaluate the proposed framework, this study selected the BioASQ rare disease medical question answering dataset as the evaluation benchmark. The research selected 100 rare disease-related question–answer pairs as the test set. For evaluation, two representative LLMs were used: \emph{DeepSeek-R1-Distill-Qwen-32B} and \emph{Qwen/QwQ-32B}, with three control configurations:
\begin{itemize}
  \item \textbf{Baseline Model (No Mechanism)}: The original model without knowledge enhancement;
  \item \textbf{Knowledge Graph + Traditional Methods}: Applying basic knowledge graphs and traditional single-dimension activation mechanisms;
  \item \textbf{Our Framework}: Including multi-granularity matching strategies, concept diversity control, fallback mechanisms, and adaptive sparse control.
\end{itemize}
All experiments were conducted in the same computing environment (NVIDIA A100 GPU). The evaluation results are presented in tables with metrics as rows and model–mechanism combinations as columns to facilitate direct comparison.

\subsection{Results and Analysis}
Model performance was evaluated using both automatic metrics and expert manual evaluation. The evaluation metrics include: BLEU (Bilingual Evaluation Understudy), which measures n-gram precision between model outputs and reference texts; ROUGE (Recall-Oriented Understudy for Gisting Evaluation), which assesses recall-oriented text similarity; \emph{Precision}, which represents the proportion of correctly identified rare diseases among all identified diseases; \emph{Recall}, which measures the proportion of actual rare diseases correctly identified by the model; and \emph{Accuracy}, which evaluates the overall correctness of diagnostic decisions.

\begin{table}[h]
\centering
\caption{Model Performance on BLEU and ROUGE Metrics}
\label{tab:bleu_rouge_pdf}
\begin{tabular}{lcccccc}
\toprule
\multirow{2}{*}{Metric} & \multicolumn{3}{c}{DeepSeek-R1-Distill-Qwen-32B} & \multicolumn{3}{c}{Qwen/QwQ-32B} \\
\cmidrule(lr){2-4} \cmidrule(lr){5-7}
& Base & Traditional & Complete & Base & Traditional & Complete \\
\midrule
BLEU-1  & 0.16 & 0.25 & 0.27 & 0.31 & 0.35 & 0.38 \\
BLEU-2  & 0.09 & 0.16 & 0.17 & 0.20 & 0.24 & 0.26 \\
BLEU-L  & 0.16 & 0.23 & 0.25 & 0.28 & 0.29 & 0.32 \\
ROUGE-1 & 0.24 & 0.31 & 0.31 & 0.35 & 0.36 & 0.38 \\
ROUGE-2 & 0.08 & 0.11 & 0.11 & 0.14 & 0.15 & 0.15 \\
ROUGE-L & 0.22 & 0.28 & 0.29 & 0.31 & 0.33 & 0.34 \\
\bottomrule
\end{tabular}
\end{table}

For the \emph{DeepSeek-R1-Distill-Qwen-32B} model, after applying our framework, BLEU-1 increased from 0.16 to 0.27 (68.8\% growth), and ROUGE-1 increased from 0.24 to 0.31 (29.2\% growth), with other metrics also showing significant improvements, as shown in Table~\ref{tab:bleu_rouge_pdf}. For \emph{Qwen/QwQ-32B}, BLEU-1 increased from 0.31 to 0.38 and ROUGE-1 from 0.35 to 0.38. These results indicate that models with stronger baseline performance can better utilize external knowledge enhancement mechanisms for diagnostic reasoning.

\begin{table}[h]
\centering
\caption{Model Performance on Precision, Recall, and Accuracy Metrics}
\label{tab:precision_recall_pdf}
\begin{tabular}{lcccccc}
\toprule
\multirow{2}{*}{Metric} & \multicolumn{3}{c}{DeepSeek-R1-Distill-Qwen-32B} & \multicolumn{3}{c}{Qwen/QwQ-32B} \\
\cmidrule(lr){2-4} \cmidrule(lr){5-7}
& Base & Traditional & Complete & Base & Traditional & Complete \\
\midrule
Precision & 0.64 & 0.71 & 0.74 & 0.78 & 0.77 & 0.80 \\
Recall    & 0.61 & 0.68 & 0.71 & 0.75 & 0.75 & 0.78 \\
Accuracy  & 0.61 & 0.76 & 0.83 & 0.87 & 0.88 & 0.89 \\
\bottomrule
\end{tabular}
\end{table}

After applying our framework to \emph{DeepSeek-R1-Distill-Qwen-32B}, precision increased from 0.64 to 0.74, recall from 0.61 to 0.71, and diagnostic accuracy from 0.61 to 0.83 (36.1\% improvement), as detailed in Table~\ref{tab:precision_recall_pdf}. The \emph{Qwen/QwQ-32B} model’s diagnostic accuracy reached 0.89, approaching the 0.90 threshold required for clinical applications. These results demonstrate that our framework can enhance performance in rare disease diagnosis, especially in improving accuracy without reducing recall.

\subsection{Manual Evaluation Results}
The research adopted the Quality Evaluation System for Text (QUEST) framework for systematic evaluation~\cite{quest2024}, inviting 12 authoritative experts in the rare-disease field to conduct manual evaluation of model-generated content. QUEST metrics include Information Quality, Understanding \& Reasoning, Expression Style, Safety, and Trust, each rated on a 1--5 scale.

\begin{table}[h]
\centering
\caption{Model Scores on QUEST Framework Manual Evaluation Metrics (1--5 points)}
\label{tab:quest_pdf}
\begin{tabular}{lcccccc}
\toprule
\multirow{2}{*}{Metric} & \multicolumn{3}{c}{DeepSeek-R1-Distill-Qwen-32B} & \multicolumn{3}{c}{Qwen/QwQ-32B} \\
\cmidrule(lr){2-4} \cmidrule(lr){5-7}
& Base & Trad. & Comp. & Base & Trad. & Comp. \\
\midrule
Information Quality        & 3.2 & 3.8 & 4.1 & 3.8 & 4.1 & 4.4 \\
Understanding \& Reasoning & 2.9 & 3.5 & 3.8 & 3.7 & 3.9 & 4.4 \\
Expression Style           & 3.4 & 3.9 & 4.2 & 4.0 & 4.2 & 4.5 \\
Safety                     & 3.6 & 4.2 & 4.5 & 4.2 & 4.4 & 4.6 \\
Trust                      & 3.0 & 3.7 & 4.0 & 3.7 & 4.0 & 4.5 \\
\bottomrule
\end{tabular}
\end{table}

For \emph{DeepSeek-R1-Distill-Qwen-32B}, after applying our framework, the information quality score increased to 4.1, understanding and reasoning to 3.8, and other dimensions also showed significant improvements~\cite{deeplearning2022,leveraging2017}. The \emph{Qwen/QwQ-32B} model with our framework reached or exceeded 4.4 points in all dimensions, indicating that its generated content is approaching specialist-physician level and has important clinical value for precise diagnosis of rare diseases~\cite{exploring2019}. 

\begin{figure}[h]
\centering
\includegraphics[width=0.9\textwidth]{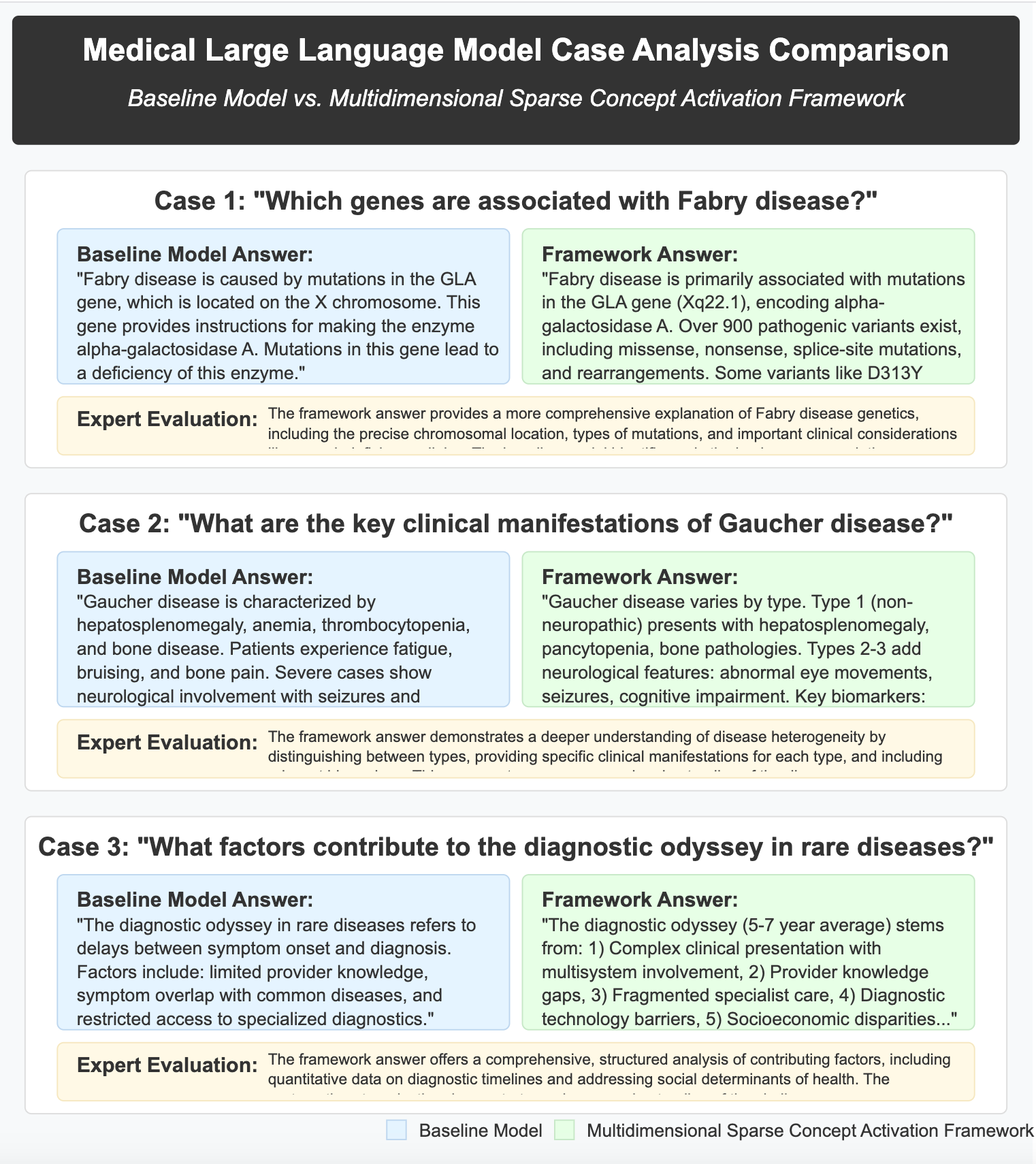}
\caption{Comparison of Model Rare Disease Answers Under Different Mechanisms}
\label{fig:comparison}
\end{figure}

Figure~\ref{fig:comparison} shows that rare-disease diagnostic answers generated with our framework exhibit significant improvements in accuracy, completeness, and professionalism. Baseline model answers often have problems such as incomplete symptom descriptions and insufficient diagnostic basis, while answers generated by our framework include comprehensive symptom descriptions, diagnostic reasoning processes, treatment recommendations, and prognosis assessments conforming to clinical norms. These advantages are further validated through typical case analysis~\cite{abdullahi2024}. This is consistent with the improvements in ``information quality'' and ``understanding and reasoning'' metrics in Table~\ref{tab:quest_pdf}, while the important role of real-world evidence in rare-disease treatment evaluation~\cite{zhou2024} also confirms that the professional expression style of our framework model approaches specialist-physician level, a conclusion supported by Gao et al.~\cite{gao2025}.

\section{Conclusion}
\label{sec:conclusion}
This research presents a multi-granularity concept sparse activation and hierarchical knowledge graph fusion framework for rare disease diagnosis. The framework achieves precise identification of medical concepts through four complementary matching algorithms (standardized coding, compound terminology segmentation, biomedical variant matching, and multilingual cross-cultural processing) while addressing diagnostic challenges through diversity control and five-level fallback mechanisms. 

Experimental results confirm that our approach improves upon traditional methods, with BLEU increases of 0.09, ROUGE increases of 0.05, and diagnostic accuracy improvements of 0.12, bringing peak model performance (0.89) close to the clinical application threshold (0.90). Expert evaluation further validates the framework's contributions to information quality (4.4/5), reasoning ability (4.4/5), and professional expression (4.5/5), with specialists confirming its potential to reduce the prolonged ``diagnostic odyssey'' experienced by rare-disease patients. Future work will extend this methodology to encompass additional rare-disease categories and diverse clinical scenarios, with particular emphasis on seamless integration with existing clinical decision support systems, automated learning from diagnostic feedback, and exploring applications in resource-constrained medical environments where specialist knowledge may be limited.

\section*{Acknowledgments}
We thank LetPub (www.letpub.com.cn) for its linguistic assistance during the preparation of this manuscript.

\section*{Declarations}
\paragraph{Funding.}
This work was supported by the Special Fund for the Central Government to Guide Local Science (Grant No. 202407AB110003), the Key Research and Development Program of Yunnan Province (Grant No. 202402AA310056), the National Natural Science Foundation of China (Grant No. 62366057), and the Scientific Research Fund Project of the Yunnan Provincial Department of Education (Grant No. 2024J0024).

\paragraph{Conflicts of interest.}
The authors declare that they have no competing interests.

\paragraph{Ethics approval.}
This study presents a theoretical framework for rare-disease diagnosis using computational methods. The research did not involve human participants, human tissue samples, or animals. All experiments were conducted on publicly available datasets (BioASQ). Therefore, ethics approval was not required. All methods were carried out in accordance with relevant guidelines and regulations.

\paragraph{Data availability.}
The data supporting the findings of this study are available within the article. The BioASQ rare disease medical question answering dataset is publicly available and can be accessed via the BioASQ challenge website (\url{https://participants-area.bioasq.org/datasets/}). Researchers interested in accessing similar datasets should contact the corresponding author with appropriate institutional approvals.

\paragraph{Author contributions.}
M.Z. conducted all experiments, developed the system, and drafted the manuscript. N.Z. and J.Q. supervised the research and provided critical guidance. G.Y. contributed to algorithm design and performance optimization. R.T. carried out data processing and analysis. All authors reviewed and approved the final manuscript.


\end{document}